\documentclass{bmvc2k}

\usepackage{algorithm}
\usepackage{algpseudocode}
\usepackage{booktabs}
\usepackage{algorithm}
\usepackage{algpseudocode}
\usepackage{booktabs}
\usepackage{multirow}

\usepackage{pifont}
\usepackage{graphicx}
\newcommand{\cmark}{\ding{51}}
\newcommand{\xmark}{\ding{55}}

\usepackage{booktabs}      
\usepackage{threeparttable}

\title{PruneNAS: Constraint Aware Neural Architecture Search based Pruning}

\addauthor{Sudhakar Sah}{sudhakar.sah@st.com}{1}
\addauthor{Nikhil Chhabra}{nikhil.chhabra@st.com}{1}
\addauthor{Honnesh Rohmetra}{honnesh.rohmetra@st.com}{1}
\addauthor{Andy Li}{andy.li@st.com}{1}
\addinstitution{
 STMicroelectronics, Inc.\\
 Toronto, Canada
}

\runninghead{Sah, Chhabra, Rohmetra, Li}{PruneNAS}

\begin{document}

\maketitle
\begin{abstract}
Neural Architecture Search (NAS) has achieved state-of-the-art performance across diverse tasks, but its high computational cost remains a major barrier. We introduce PruneNAS, a novel framework that integrates structured pruning into NAS to efficiently explore high-performing architectures while leveraging pretrained models. Unlike conventional NAS methods that train each candidate from scratch, PruneNAS generates a candidate pool by pruning pretrained networks at multiple granularity levels and formulates the selection as a global optimization problem using training-free importance metrics. This approach enables rapid evaluation of candidate architectures under strict resource constraints, dramatically reducing search time from hundreds of GPU hours to mere minutes. Experiments on standard benchmarks demonstrate that PruneNAS discovers compact, high-accuracy architectures that nearly halve the computational cost (MACs) of prior methods on ResNet-50, while also showing strong generalization on object detection tasks with YOLO. Our method bridges structured pruning and NAS, offering a practical solution for resource-constrained model deployment.
\end{abstract}

\section{Introduction\label{sec:introduction}}
Deep neural networks have achieved remarkable success across diverse computer vision tasks. However, the increasing scale of these models imposes significant computational burdens, which complicates their deployment on resource-constrained edge devices \cite{chen2019deep, menghani2023efficient}. To resolve this, channel pruning \cite{you2019gate, li2016pruning, hou2022chex, ding2019centripetal} has emerged as a highly effective technique. By structurally removing entire channels and their associated weights, it directly reduces the model's memory footprint, computational cost and accelerates inference. 

Determining which channels to remove is commonly guided by importance criteria \cite{yu2022hessian, he2017channel, molchanov2019importance}.
These criteria estimate each channel's importance to the network output and are used to rank and remove channels under a target resource budget.
However, identifying the optimal distribution of pruned channels across layers such that a strict hardware resource constraint is met exactly remains a challenge, as it is typically handled through some pre-defined heuristics.

Recent graph-based methods~\cite{fang2023depgraph} address layer dependencies but typically lack sound non-uniform pruning strategies across groups, ignoring that different structural components have different sensitivities to pruning~\cite{he2018amc, fpgm}.  Neural Architecture Search (NAS) often unifies search and pruning to find non-uniform distributions via searching for optimal structures within a search space. However, a brute-force approach requires the combinatorial exploration of all channel configurations, making it computationally impossible. While recent approaches~\cite{cai2019once, meta, autoslim} mitigate this by utilizing weight-sharing super networks or proxy evaluators, they often shift the computational bottleneck, requiring extensive training costs and GPU hours before the actual architecture search begins. As summarized in Table~\ref{tab:efficiency_comparison}, these methods require hundreds to thousands of GPU hours simply to build the search space. Furthermore, while zero-cost proxies~\cite{lee2018snip, Wang2020Picking, mellor2021neural,tanaka2020pruning,lee2024az} offer efficient evaluation without full training, their application has largely been restricted to ranking entire architectures rather than selecting structured components under strict global constraints.

\begin{table*}[t]
\centering
\caption{Comparison of state-of-the-art methods on ResNet50/MobileNet and ImageNet.}
\label{tab:efficiency_comparison}
\resizebox{\textwidth}{!}{
\begin{tabular}{l c c c c l}
\toprule
\textbf{Method} & \textbf{Training-Free} & \textbf{Combinatorial} & \textbf{Pre-Search} & \textbf{Approx. GPU Hours} & \textbf{Search} \\
& \textbf{Prep} & \textbf{Search} & \textbf{Product} & & \textbf{Mechanism} \\
\midrule
OFA \cite{cai2019once} & \xmark & \cmark & Massive supernet & $\sim$1,200 & Evolutionary \\
AutoSlim \cite{autoslim} & \xmark & \xmark & Slimmable Net & $>$300 & Greedy \\
MetaPruning \cite{meta} & \xmark & \cmark & PruningNet & $>$200 & Evolutionary \\
DepGraph \cite{fang2023depgraph} & \cmark & \xmark & None (Standard Pruning) & NA & Heuristic \\
\textbf{PruneNAS (Ours)} & \textbf{\cmark} & \textbf{\cmark} & \textbf{Lookup tables} & \textbf{minutes} & \textbf{ILP} \\
\bottomrule
\end{tabular}
}
\end{table*}

\footnotetext[1]{The authors of OFA report ~1,200 GPU hours on V100s before search. AutoSlim and MetaPruning hours are conservatively estimated based on standard ResNet-50 ImageNet training (~200 GPU hours per 90 epochs). Slimmable networks use a computationally heavy "sandwich rule" \cite{yu2019universally}, making its 50-epoch preparation equivalent to ~200 standard epochs. MetaPruning trains a PruningNet for $\sim$22 epochs, but incurs significant per-iteration overhead from meta-gradient backpropagation.}
\footnotetext[2] {See Appendix (Table \ref{tab:granularity}) for detailed wall-clock benchmarks.}



To bridge this gap, we introduce PruneNAS, a framework that formulates channel allocation as a global constrained optimization problem. 
PruneNAS searches per-group channel widths within the pretrained model, retaining its layers and operators for native deployment on target edge hardware.
For each structural group in the network, a discrete set of pruned candidates is scored using an importance criterion and stored in accuracy and resource lookup tables without any training. An Integer Linear Programming (ILP) solver then selects the globally optimal per-group channel configuration under a target resource budget. Finally, the same importance criterion used during search is used to prune the channels. Unlike other NAS methods, we skip the need for pre-search training entirely. This reduces architecture discovery time from potentially hundreds of GPU hours to mere minutes. Because the tables are computed once, the solver can be re-run for any budget instantly, making it straightforward to generate a family of compressed models across different deployment budgets from a single pass. We make the following contributions:

\begin{itemize}
    \item We formulate channel pruning as a constrained optimization problem using a decoupled tabulation strategy. Our approach eliminates the computationally expensive need for training candidate models for the search phase.
    \item We present a highly versatile pruning framework that supports various importance criteria (e.g., magnitude, Taylor, etc) and accommodates multiple resource constraints, including Multiply–Accumulate operations (MACs) and the number of parameters.
    \item We demonstrate through empirical experiments that PruneNAS achieves SOTA accuracy to resource trade-offs on vision classification tasks, outperforming existing methods by up to 1\% on ResNet-50 \cite{he2016deep} and ImageNet-1k \cite{deng2009imagenet}.
\end{itemize}

\section{Related Work}
\label{sec:related_work}

\subsection{Importance and Zero-Cost Estimation}

Structured pruning reduces the parameters in a network by removing entire groups such as neurons or channels~\cite{wen2016learning, he2017channel, fang2023depgraph}. Traditionally, this is achieved by analyzing the converged weights of pre-trained models with an importance criterion that determines what to keep. Magnitude-based pruning prunes channels with smaller $\ell_1$-norm~\cite{li2017pruning}. Taylor expansion estimates the impact of pruning on the loss by computing the product of activations and their gradients~\cite{molchanov2017pruningtaylor, molchanov2019importance}. Second-order methods utilize Hessian-based approximations sensitivity~\cite{lecun1989optimal, yu2022hessian} and Fisher information \cite{theis2018faster, liu2021group} to evaluate parameter. 
Layer Adaptive Magnitude-based Pruning (LAMP) \cite{lee2020layer} improves magnitude pruning by using a rescaled score to automatically allocate sparsity, preventing layer collapse without manual tuning. Filter Pruning via Geometric Median (FPGM) \cite{fpgm} prunes channels that are closest to the layer's geometric median, focusing on removing redundancies rather than magnitude. Interestingly, random pruning has been shown to perform competitively in some settings, questioning the necessity of complex criteria~\cite{liu2022unreasonable}.

Parallel to these, zero-cost proxies (ZCPs) estimate network quality without training. Representative methods include SNIP, GraSP, SynFlow, NASWOT, Fisher, Zico, and AZ-NAS, which use gradients, activation patterns, or parameter statistics to rank candidate architectures~\cite{lee2018snip, Wang2020Picking, tanaka2020pruning, mellor2021neural, abdelfattah2021zero, li2023zico, lee2024az}. Although ZCPs have been widely studied in NAS, their application to structured pruning remains limited. Existing methods primarily focus on ranking entire, randomly initialized architectures rather than selecting specific structured components under strict global constraints. Furthermore, as we demonstrate in our empirical analysis, traditional untrained ZCPs can sometimes yield noisy signals when scaling channel search space (as shown in ablation). Bridging this gap is essential for a reliable training-free structured pruning framework. PruneNAS adopts the computational efficiency of ZCP but applies it via the more robust importance criteria on pre-trained weights. This enables a smooth and more reliable tabulation process for our search algorithm.

\subsection{Dependency Graph}

Graph-based structured pruning methods have recently gained attention for their ability to handle inter-layer dependencies in modern neural networks. In particular, DepGraph~\cite{fang2023depgraph} models structural relationships as a dependency graph, enabling consistent pruning across operations such as residual connections and concatenations. This addresses a key limitation of earlier layer-wise pruning approaches~\cite{li2016pruning, molchanov2017pruningtaylor}, which often ignore cross-layer constraints and may produce suboptimal or invalid architectures. By explicitly encoding these dependencies, DepGraph enables globally consistent pruning decisions while preserving network integrity. However, existing graph-based methods typically apply uniform pruning strategies across groups and rely on predefined importance criteria. In contrast, different groups may exhibit varying sensitivity to pruning, and our approach exploits this by adaptively selecting pruning levels for each group under global resource constraints.
Approaches based on soft masking and budget-aware regularization~\cite{lemaire2018structured} enable end-to-end learning of structured sparsity under constraints, but often introduce additional optimization complexity.

\subsection{Neural Architecture Search and Structured Pruning}

Structured pruning is closely related to neural architecture search (NAS), with recent works increasingly formulating pruning as a search over channel configurations. One-Shot NAS methods such as AutoSlim \cite{autoslim}, MetaPruning \cite{meta}, Once-for-All \cite{cai2019once}, and Slimmable Networks \cite{slimmable} navigate this combinatorial space by training weight-sharing super-networks or meta-networks from which sub-networks are subsequently extracted.

Differentiable approaches relax channel selection and jointly optimize pruning decisions and weights~\cite{liu2019darts}, while ProxylessNAS \cite{cai2018proxylessnas} and FBNet \cite{wu2019fbnet} incorporate hardware metrics such as latency into the search objective. Pruning-as-Search (PaS) \cite{li2022pas}, TAS \cite{dong2019network}, and NAP \cite{ding2022nap} similarly cast pruning as an architecture-search problem using gradient-based or DARTS-style optimization.

Many NAS methods additionally search over network structure or operators, which can complicate deployment on edge accelerators with limited operator support. LANA~\cite{molchanov2022lana} replaces each layer with an operation drawn from a rich candidate pool, APQ~\cite{wang2020apq} jointly searches architecture, pruning, and quantization, and CONNAS~\cite{deruyter2026connas} extends ILP-based NAS to multiple hardware objectives. Training-free methods such as TTNAS~\cite{lin2024ttnas} and PATNAS~\cite{yang2025patnas} still search over generalized supernetworks or randomly initialized architectures, while HyperNAS~\cite{lv2026hypernas} contributes a performance predictor rather than a search algorithm. Closest to our work, PaS~\cite{li2022pas} also searches channel configurations, but learns its pruning policy by introducing depthwise-convolution layers; in contrast, PruneNAS searches the structured pruning space of a fixed pretrained model using importance scores and preserves the original operators for edge deployment.


\section{Method\label{sec:method}}
\begin{figure}[t] 
    \centering
    \includegraphics[width=\linewidth]{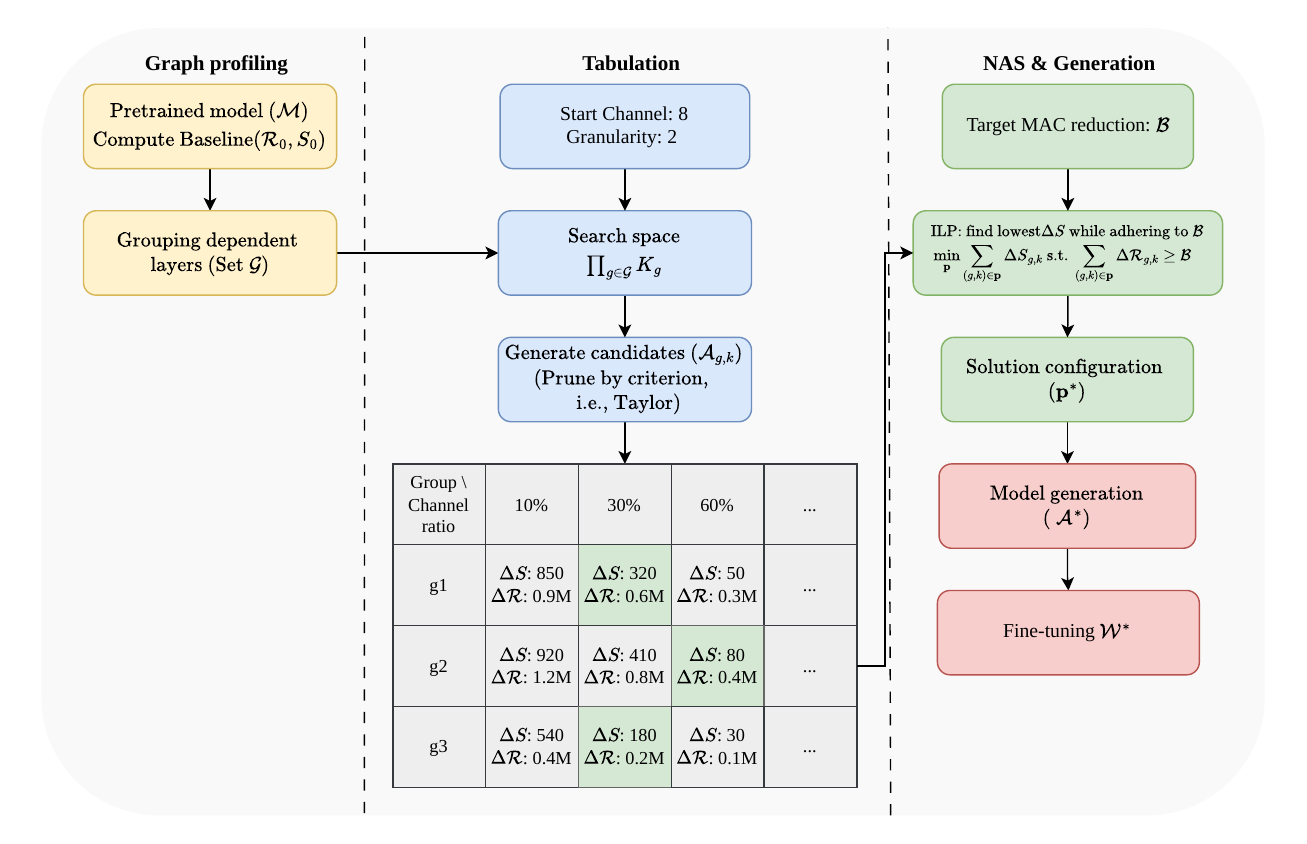} 
    \caption{The PruneNAS pipeline. (Left) A dense pretrained model is parsed into structurally coupled channel groups. (Center) A local search space is tabulated using zero-cost proxies to estimate resource reductions ($\Delta \mathcal{R}$) and importance tradeoffs ($\Delta S$). (Right) An ILP solver selects the globally optimal configuration under target resource budgets, producing a pruned architecture ready for fine-tuning.}
    \label{fig:pipeline}
\end{figure}

We propose a group-wise pruning-aware neural architecture search strategy that decomposes the global search problem into a training-free tabulation phase followed by a global constrained optimization step. 
As illustrated in Figure~\ref{fig:pipeline}, our method integrates a structured search space defined over layer groups, where each group comprises multiple layers and the search variables correspond to the number of channels retained or pruned within each group, together with zero-cost proxy tabulation and resource-aware constraint handling into a cohesive pipeline.

Following a preliminary graph profiling step to partition the pretrained model into structurally valid pruning groups, the core of our approach centers on two primary phases. First, in the tabulation phase, we generate candidate networks by pruning one group at a time and estimate their resource reductions and accuracy trade-offs using an importance metric (e.g., Taylor). These candidates along with their KPIs are stored in lookup tables. Next, during the NAS and generation phase, we formulate the global architecture search as an ILP (PuLP package~\cite{dunning2011pulp}) problem, which explores the tabulated space to find the globally optimal configuration that minimizes the total importance score drop while strictly conforming to a target resource budget. The resulting architecture is then constructed and fine-tuned to recover task performance.

\subsection{Graph Profiling}
\begin{figure}
    \centering
    \includegraphics[width=0.6\linewidth]{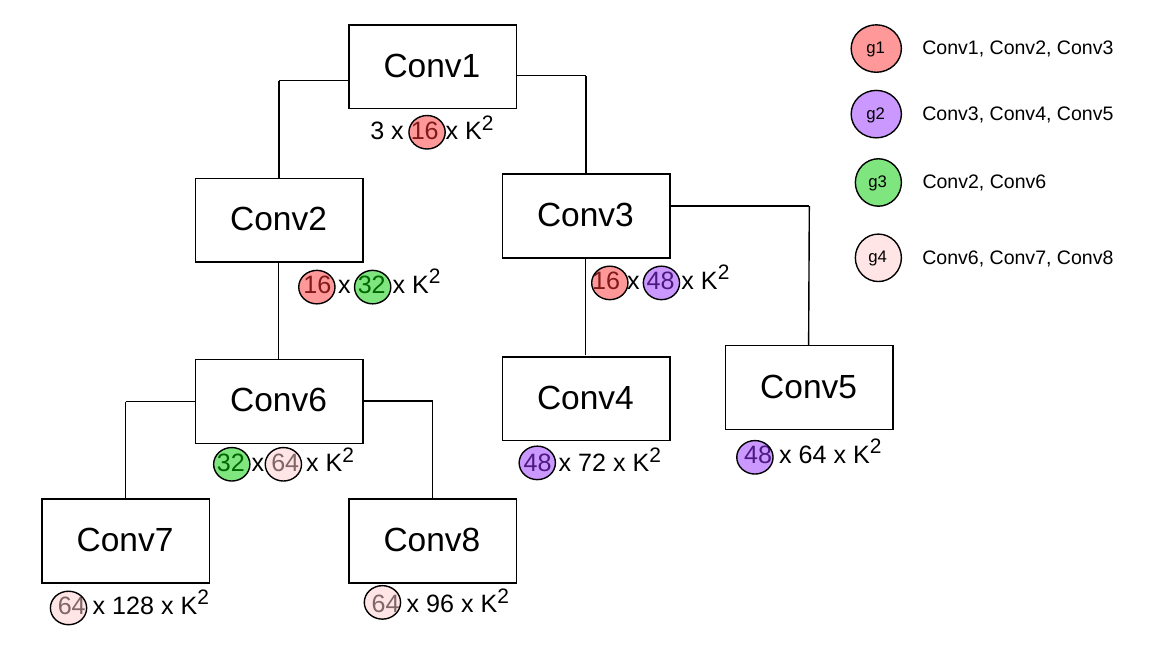}
    \caption{Independent groups in a sample model based on common number of input and output channels}
    \label{fig:dependent_groups}
\end{figure}

Let a dense, pre-trained network be parameterized by $\mathcal{W}$. To ensure structural integrity and valid tensor operations during pruning, we partition $\mathcal{W}$ into $N$ groups $\mathcal{G} = \{g_1, g_2, ... g_N\}$ as shown in figure ~\ref{fig:dependent_groups}. For each group, we define a discrete set of candidates $\mathcal{C}_i = \{\mathcal{C}_{i,1}, \mathcal{C}_{i,2}, ... \mathcal{C}_{i, M} \}$ corresponding to the group's remaining channels after pruning. The objective of our method is to select exactly one optimal configuration $c_{i}^* \in C_i$ for every group to construct the final solution, such that the total importance score loss (i.e., cumulative accuracy degradation) is minimized, subject to achieving a resource reduction of at least $\mathcal{B}$.

To ensure structural consistency, we partition the network into a set of pruning groups $\mathcal{G}$, where each group contains layers with pruning dependencies (e.g., due to residual connections or channel alignment constraints). Channels within a group are pruned jointly as proposed by Fang et. al.~\cite{fang2023depgraph}.

\subsection{Candidate Generation}

For each group $g \in \mathcal{G}$, we generate candidate architectures by pruning a small number of channels according to a predefined granularity. Specifically, for each pruning level $k$, we construct a candidate model $\mathcal{A}_{g,k}$ by removing $k$ channels from group $g$ while keeping all other groups unchanged. 

The tabulation stage evaluates only the group-wise pruning variations independently, resulting in $\sum_{g \in \mathcal{G}} K_g$ measured candidates. The exhaustive architecture space is then implicitly defined by the Cartesian product of these group-level choices, which the ILP solver optimizes over without explicit enumeration:

\begin{equation}
|\mathcal{T}_{\text{full}}| = \prod_{g \in \mathcal{G}} K_g
\end{equation}

Where each candidate architecture is represented as:

\begin{equation}
\mathcal{A} = \{c_1, c_2, \dots, c_g\}
\end{equation}
where $c_g$ denotes the number of retained channels in group $g$.

\subsection{Tabulation}
Each candidate model $\mathcal{A}_{g,k}$ is evaluated to compute:
\begin{itemize}
    \item Resource metrics $\mathcal{R}_{g,k}$ (FLOPs, MACs, latency, power, RAM etc.)
    \item Estimated accuracy score $S_{g,k}$ using an importance-based zero-cost proxy
\end{itemize}

We further compute the relative changes with respect to the baseline. And these delta values form the entries of the accuracy and resource tables:

\begin{equation}
\Delta \mathcal{R}_{g,k} = \mathcal{R}_0 - \mathcal{R}_{g,k}  ,\quad
\Delta S_{g,k} = S_0 -  S_{g,k}
\end{equation}

\subsubsection{Resource Tabulation}
Resource tabulation is performed using metrics such as FLOPs, MACs, parameter count, latency,and memory (RAM) usage. Resource estimation is performed via profiling tools.
The baseline model is used to establish the reference resource value $R_o$. We then construct the resource table by storing the delta resource values for all variations within each group relative to this baseline. For hardware aware tabulation, the model is executed on actual hardware to gather the latency, RAM, Flash and power budget, whereas hardware agnostic calculations use proxies such as latency, RAM and power.   




\subsubsection{Accuracy Tabulation}

To efficiently estimate the baseline score $S_0$ and the candidate scores $S_{g,k}$ without the cost of fine-tuning, we explored the option to use zero-cost proxies (ZCP) methods such as SNIP~\cite{lee2018snip} and GraSP~\cite{Wang2020Picking} relying on gradient information, which approach zero at convergence for pretrained models and give noisy importance evaluation. Similarly, SynFLOW~\cite{tanaka2020pruning} is independent of training data, and does not make use of trained models. After careful comparative analysis as explained quantitatively in section~\ref{sec:accuracy_estimators}, we adopted importance criteria such as Hessian, ~\cite{yu2022hessian}, $L_1$ weight magnitude, Taylor expansion~\cite{molchanov2017pruningtaylor} as a proxy to estimate the impact of pruning on performance of the model. We use the same method as an accuracy estimator during the search phase to approximate the candidates as well as for channel pruning once the optimal configuration is finalized. 

Let $W_c$ denote the set of weights associated with a specific channel $c$. We leverage channel-wise importance scores derived from the learned representations of a pretrained model. Specifically, the change in accuracy (delta accuracy) for a candidate is estimated as the sum of the importance values of the removed channels. For a single layer's channel $c$, the metrics are mathematically expressed as follows:

\begin{algorithm}[!htb]
\caption{PruneNAS: Group-wise Candidate Generation and Search}
\label{alg:algorithm}
\begin{algorithmic}[1]
\Require Pretrained model $\mathcal{M}$, pruning groups $\mathcal{G}$, resource budget $\mathcal{B}$
\Ensure Optimal pruning configuration $\mathbf{p}^*$

\State Initialize baseline architecture $\mathcal{A}_0 \leftarrow \mathcal{M}$
\State Compute baseline resources $\mathcal{R}_0$ and baseline score $S_0$

\State $\mathcal{T} \leftarrow \emptyset$ \Comment{Candidate table}

\For{each group $g \in \mathcal{G}$}
    \State Determine pruning granularity levels $\{k_1, k_2, \dots\}$ for group $g$
    
    \For{each pruning level $k$ in group $g$}
        \State $\mathcal{A}_{g,k} \leftarrow$ prune $k$ channels from group $g$ in $\mathcal{A}_0$
        
        \State Compute resource $\mathcal{R}_{g,k}$ (FLOPs, MACs, latency, etc.)
        \State Estimate score $S_{g,k}$ using importance-based zero-cost proxy
        
        \State Compute deltas:
        \[
        \Delta \mathcal{R}_{g,k} = \mathcal{R}_0 - \mathcal{R}_{g,k}  ,\quad
        \Delta S_{g,k} = S_0 -  S_{g,k}
        \]
        
        \State Store $(g, k, \mathcal{R}_{g,k}, S_{g,k}, \Delta \mathcal{R}_{g,k}, \Delta S_{g,k})$ in $\mathcal{T}$
    \EndFor
\EndFor

\Statex

\Comment{Search optimal combination under constraints}
\State $\mathbf{p}^* \leftarrow \emptyset$

\State Solve:
\[
\min_{\mathbf{p}} \quad \sum_{(g,k) \in \mathbf{p}} \Delta S_{g,k}
\quad \text{s.t.} \quad
\sum_{(g,k) \in \mathbf{p}} \Delta \mathcal{R}_{g,k} \geq \mathcal{B}
\]

\Statex

\State Construct final architecture $\mathcal{A}^*$ using $\mathbf{p}^*$

\State \Return $\mathcal{A}^*$

\end{algorithmic}
\end{algorithm}
\subsection{Constrained Optimization}
Given the tabulated candidates, we formulate the global search as a constrained optimization problem that selects one pruning action per group:
\begin{equation}
\label{eq:ilp}
\min_{\mathbf{p}} \quad \sum_{(g,k) \in \mathbf{p}} \Delta S_{g,k}
\quad \text{s.t.} \quad
\sum_{(g,k) \in \mathbf{p}} \Delta \mathcal{R}_{g,k} \geq \mathcal{B}
\end{equation}
where $\mathbf{p}$ denotes the set of selected pruning actions and $\mathcal{B}$ is the resource budget. 

The formulation naturally extends to multiple resource constraints. For example, when jointly optimizing MACs and Flash memory, we define separate reduction terms $\Delta \mathcal{R}_{g,k}^{\mathrm{MAC}}$ and 
$\Delta \mathcal{R}_{g,k}^{\mathrm{flash}}$ with corresponding target reductions $\mathcal{B}_{1}$ and $\mathcal{B}_{2}$.

 \begin{equation}
\min_{\mathbf{p}} \quad \sum_{(g,k) \in \mathbf{p}} \Delta S_{g,k}
\quad \text{s.t.} \quad
\sum_{(g,k) \in \mathbf{p}} \Delta \mathcal{R}_{g,k}^{\mathrm{MAC}} \geq \mathcal{B}_{1}
\end{equation}
 \begin{equation}
\min_{\mathbf{p}} \quad \sum_{(g,k) \in \mathbf{p}} \Delta S_{g,k}
\quad \text{s.t.} \quad
\sum_{(g,k) \in \mathbf{p}} \Delta \mathcal{R}_{g,k}^{\mathrm{flash}} \geq \mathcal{B}_{2}
\end{equation}
In practice, we solve the optimization independently for each resource target (e.g., MACs and Flash). The final pruning configuration is then obtained by taking, for every group, the minimum retained channels across the resulting solutions, ensuring that both constraints are satisfied conservatively. The complete procedure is summarized in Algorithm \ref{alg:algorithm}.

\subsection{Finetuning and Training}
The optimal solution $\mathbf{p}^*$ specifies the number of channels to prune for each group. Using $\mathbf{p}^*$, we construct the final pruned architecture $\mathcal{A}^*$ and train it to completion. Our method gives the flexibility to prune either one-shot or iteratively:
\begin{itemize}
    \item The model is pruned to the target $\mathcal{A}^*$ at once, followed by standard fine-tuning to completion.
    \item Iterative Fine-Tuning: The pruning process is divided into $n$ steps, removing $p_g^* / n$ channels per group at each step. The intermediate network is fine-tuned for $m$ epochs after each increment over $n \times m$ total epochs.
\end{itemize}

\section{Experiments\label{sec:experiment}}

We conduct extensive experiments on ImageNet-1k \cite{deng2009imagenet} and Flowers102 dataset\cite{nilsback2008automated}. We first benchmark PruneNAS against state-of-the-art structured pruning methods on the large-scale ImageNet-1k dataset using ResNet-50 \cite{he2016deep} and MobileNetV2 \cite{sandler2018mobilenetv2} to demonstrate its competitive accuracy to resource trade-offs. Next, we use the Flowers102 dataset with ResNet-18 and MobileNetV2 to compare our one-shot and iterative pruning variants. We then validate the practical deployability of our architectures through on-device hardware benchmarking. Furthermore, we evaluate the framework's cross-task generalization to YOLO object detection architectures.

Candidate tabulation and global search processes were primarily conducted on NVIDIA A100/A6000 GPUs, taking only a few minutes per search due to the training-free nature of our proxy method. For all fine-tuning, we use the default training hyperparameters from the timm library \cite{wightman2021resnet} (300 epochs, warmup epochs-5,  SGD optimizer, momentum 0.9, weight decay -1e-4, LR schedule- cosine decay, batch size - 512). Our iterative variant does not increase the training budget: the epochs consumed across all intermediate pruning rounds and the final fine-tuning sum to the same total as the one-shot variant.

\subsection{Classification using ImageNet}

\begin{table*}[t]
\centering
\caption{Pruning results on ImageNet. $\Delta$ Acc. denotes the accuracy drop after pruning. MACs are in billions (G) and Params in millions (M). A group-wise prune ratio ceiling of 70\% is applied to MobileNetV2 and no ceiling is applied to ResNet-50. ResNet-50 was fine-tuned with one-shot and MobileNetV2 was fine-tuned iteratively for 5 rounds. (*Not a ResNet variation but model with similar GMACs) \\}
\label{tab:pruning_results}
\begin{tabular}{llccccc}
\toprule
\textbf{Arch.} & \textbf{Method} & \textbf{Base} & \textbf{Pruned} & \textbf{$\Delta$ Acc.} & \textbf{MACs (G)} & \textbf{Params (M)} \\

\midrule
ResNet-50 & Baseline & 76.15 & - & - & 4.13 & 25.6 \\
& ThiNet~\cite{thinet} & 72.88 & 72.04 & -0.84 & 2.44 & - \\
& SSS~\cite{wen2016learning} & 76.12 & 74.18 & -1.94 & 2.82 & - \\
& SFP~\cite{sfp} & 76.15 & 74.61 & -1.54 & 2.40 & - \\
& AutoSlim~\cite{autoslim} & 76.10 & 75.60 & -0.50 & 2.00 & - \\
& FPGM~\cite{fpgm} & 76.15 & 75.50 & -0.65 & 2.38 & - \\
& Taylor~\cite{molchanov2019importance} & 76.18 & 74.50 & -1.68 & 2.25 & - \\
& Slimmable~\cite{slimmable} & 76.10 & 74.90 & -1.20 & 2.30 & - \\
& CCP~\cite{ccp} & 76.15 & 75.50 & -0.65 & 2.11 & - \\
& AOFP-C1~\cite{aofp} & 75.34 & 75.63 & +0.29 & 2.58 & - \\
& TAS~\cite{tas} & 77.46 & 76.20 & -1.26 & 2.31 & - \\
& GFP~\cite{gfp} & 76.79 & 76.42 & -0.37 & 2.04 & - \\
& GReg-2~\cite{greg} & 76.13 & 75.36 & -0.77 & 2.77 & - \\
& PaS-C*~\cite{li2022pas} & - & {75.50} & {-} & {2.40} & {-} \\
& DepGraph~\cite{fang2023depgraph} & 76.15 & {73.58} & {-2.57} & {1.37} & {8.77} \\
& \textbf{PruneNAS} & 76.15 & \textbf{74.54} & \textbf{-1.61} & \textbf{1.37} & {8.60} \\

\midrule
MobileNetV2 & Baseline & 71.87 & - & - & 0.33 & 3.5 \\
& NetAdapt~\cite{netadapt} & - & 70.00 & - & 0.24 & - \\
& Meta~\cite{meta} & 74.70 & 68.20 & -6.50 & 0.14 & - \\
& GFP~\cite{gfp} & 75.74 & 69.16 & -6.58 & 0.15 & - \\
& DepGraph~\cite{fang2023depgraph} & 71.87 & {68.46} & {-3.41} & 0.15 & - \\
& \textbf{PruneNAS} & 71.57 & \textbf{67.80} & \textbf{-3.77} & \textbf{0.15} & {1.57} \\

\bottomrule
\end{tabular}
\end{table*}

\begin{table}[h]
\caption{YOLOv8 Results on MSCOCO~\cite{yolov8_ultralytics} dataset trained using 640px resolution\\}
\centering
\begin{tabular}{lccc}
\toprule

\textbf{Model} & \textbf{mAP@50--95} & \textbf{FLOPS (G)} & \textbf{Parameters (M)} \\
\midrule
YOLOv8s~\cite{yolov8_ultralytics} (Dense) & 44.9 & 28.60 & 11.17 \\
\midrule
YOLOv8n~\cite{yolov8_ultralytics} (Dense) & 37.3 & 8.70 & 3.20 \\
\textbf{YOLOv8s (PruneNAS)} & \textbf{38.4} & \textbf{8.48} & 2.32 \\
\textbf{YOLOv8s (PruneNAS)} & \textbf{41.4} & 15.74 & 3.28 \\

\bottomrule
\end{tabular}

\label{tab:yolov8_prunenas_results}
\end{table}

We evaluate the effectiveness of the proposed PruneNAS framework on the ImageNet dataset and compare it with a wide range of state-of-the-art structured pruning methods. Table~\ref{tab:pruning_results} reports the Top-1 accuracy before and after pruning, the corresponding accuracy drop ($\Delta$ Acc.), as well as the computational complexity in terms of MACs and model size (parameters).

\paragraph{ResNet-50.}
For ResNet-50, PruneNAS achieves a favorable trade-off between accuracy and efficiency at a comparable computational budget. Specifically, at 1.37G MACs, PruneNAS obtains a Top-1 accuracy of 74.54\%, resulting in a 1.61\% drop from the baseline. This outperforms DepGraph~\cite{fang2023depgraph}, which incurs a larger accuracy drop of 2.57\% under the same MAC constraint. Compared to other pruning approaches such as ThiNet, SFP, and FPGM, PruneNAS demonstrates competitive or superior accuracy while operating at lower or similar computational costs. Notably, PruneNAS reaches accuracy comparable to PaS-C~\cite{li2022pas} (74.54\% vs.\ 75.50\%) at nearly half the compute (1.37G vs.\ 2.40G MACs) and, unlike PaS, inserts no additional depthwise-convolution layers, preserving the original operator set for edge deployment.

\paragraph{MobileNetV2.}
On MobileNetV2 \cite{sandler2018mobilenetv2}, we further validate the effectiveness of PruneNAS under a lightweight architecture setting. At 0.15G MACs, PruneNAS achieves 67.80\% Top-1 accuracy. While this reflects a slightly more accuracy drop compared to some methods like Depgraph~\cite{fang2023depgraph}, it is important to note that PruneNAS operates under strict compression constraints and achieves a significantly reduced parameter count (1.57M). Compared to DepGraph, PruneNAS provides a more compact model with fewer parameters, highlighting its ability to aggressively compress networks. We additionally validate PruneNAS on medium-scale CIFAR-100 with ResNet-18 (Appendix, Table~\ref{tab:mac_cifar100}), where it consistently outperforms DepGraph across all MAC-reduction ratios from $1.5\times$ to $9\times$.

\subsection{Object Detection using COCO}
\paragraph{YOLOv8}
The results in Table~\ref{tab:yolov8_prunenas_results} highlight the effectiveness of PruneNAS in adapting a stronger baseline (YOLOv8s) to tighter resource budgets while retaining competitive accuracy. When pruning YOLOv8s to match the compute footprint of the lighter YOLOv8n ($\sim$8.70 FLOPS), PruneNAS achieves a clear improvement in mAP (38.4 vs. 37.3), demonstrating that structured pruning guided by search can outperform a hand-designed compact model at the same compute level. Similarly, the pruned YOLOv8s variant (3.28M parameters) preserves significantly higher accuracy (41.4 mAP) than YOLOv8n (37.3). 

Overall, PruneNAS consistently achieves strong performance across different architectures, particularly under high compression regimes. The results suggest that the proposed search-based pruning strategy is effective in exploring the trade-off between accuracy and efficiency. In particular, PruneNAS shows clear advantages over dependency-based pruning methods at identical MAC budgets, demonstrating its ability to better preserve important structures during pruning.


\begin{table}[t]
\centering
\caption{Accuracy to MAC Reduction trade-offs on the Flowers102 dataset using one-shot and iterative training. Pruned represents the accuracy immediately after pruning and before training. A maximum group-wise pruning ratio ceiling of 70\% is imposed to prevent layer collapse, except for MobileNetV2 at 9$\times$ compression, where the ceiling is relaxed to 80\%. \\}
\label{tab:mac_flowers102}
\begin{tabular}{llccccc}
\toprule
\multirow{2}{*}{\textbf{Model}} & \multirow{2}{*}{\textbf{Variation}} & \multicolumn{5}{c}{\textbf{MAC Reduction}} \\
\cmidrule(lr){3-7}
 &  & 1.0x & 1.5x & 3.0x & 6.0x & 9.0x \\
\midrule

\multirow{4}{*}{ResNet18}
 & Pruned             & 91.87 & 27.79 & 1.14 & 0.47 & 0.47 \\
 & DepGraph~\cite{fang2023depgraph}            & 91.87 & \textbf{90.32} & 86.27 & 79.87 & 76.01 \\
 & PruneNAS(\textit{one-shot})      & 91.87 & 89.40 & 85.35 & 82.55 & 74.66 \\
 & \textbf{PruneNAS}(\textit{iterative}
 )  & 91.87 & 90.13 & \textbf{87.25} & \textbf{84.89} & \textbf{82.79} \\

\midrule

\multirow{4}{*}{MobileNetV2}
 & Pruned            & 93.62 & 2.46 & 1.40 & 1.19 & 0.60 \\
 & DepGraph~\cite{fang2023depgraph}            & 93.62 & 90.86 & 86.05 & 64.63 & 62.32 \\
 & \textbf{PruneNAS}(\textit{iterative})  & 93.62 & \textbf{90.86} & \textbf{86.25} & \textbf{84.35} & \textbf{81.13}  \\

\bottomrule
\end{tabular}
\end{table}

\subsection{Ablation Studies}
\begin{table}[t]
\centering
\caption{Comparison of pruned accuracy using various importance criteria and MAC reductions, tested with ResNet-18 and the Flowers102 dataset. A max group-wise prune ratio ceiling of 70\% is applied to prevent layer collapse. To ensure statistical robustness, the performance metrics for the importance criteria ablations are averaged over 3 random seeds.  \\} 
\label{tab:sparsity_accuracy}
\begin{tabular}{lccccc}
\toprule
\multirow{2}{*}{\textbf{Importance}} & \multicolumn{5}{c}{\textbf{MAC Reduction}} \\
\cmidrule(lr){2-6}
 & \textbf{1.5$\times$} & \textbf{3$\times$} & \textbf{6$\times$} & \textbf{9$\times$} & \textbf{Avg.} \\
\midrule
Unpruned  & \multicolumn{1}{c}{91.87} \\
\midrule
Magnitude & 89.21 ± 0.35 & 84.74 ± 0.36  &  83.69 ± 0.28 &  79.77 ± 0.29& 84.35 ± 0.16  \\
Taylor    &  89.51 ± 0.27 & 86.65 ± 0.22  &  83.50 ± 0.13 & 77.03 ± 0.37 & 84.17 ± 0.13  \\
Hessian   &  89.23 ± 0.40 &  83.72 ± 0.50 &  71.13 ± 2.42 & 50.28 ± 4.05 & 73.59 ± 1.19  \\
FPGM   & 88.75 ± 0.41  & 83.37 ± 0.21  &  81.91 ± 0.25 & 77.94 ± 0.43 & 82.99 ± 0.17  \\
LAMP   & \textbf{89.97} ± 0.34  & \textbf{88.12 ± 0.35}  & \textbf{85.28 ± 0.13}  & \textbf{80.30 ± 0.06} &  \textbf{85.67 ± 0.13} \\
\bottomrule
\end{tabular}
\end{table}

\subsubsection{Accuracy Estimators}

Appendix Table~\ref{tab:estimators_ablation} presents a comparison of top-1 accuracy achieved by different zero-cost estimators under a fixed 9$\times$ MAC reduction constraint using a ResNet-18 model on the Flowers102 dataset. The dense (unpruned) baseline achieves a top-1 accuracy of 91.87\%, providing a reference for evaluating the effectiveness of each estimator under aggressive compression.

Taylor-based scoring achieves the highest top-1 accuracy of 82.79\%, indicating its superior ability to preserve salient network parameters during pruning. NWOT follows closely with 82.61\%, demonstrating competitive performance despite its data-agnostic formulation. L2-Norm achieves 82.35\%, suggesting that magnitude-based pruning remains a strong baseline even at high compression ratios. In contrast, AZ-NAS\cite{lee2024az} yields a lower accuracy of 81.80\%, highlighting comparatively weaker ranking capability under the same constraint.

Overall, the results indicate that gradient-based criteria, particularly Taylor expansion, provide more reliable importance estimation for structured pruning at high compression levels. The relatively small performance gap between methods, however, suggests that multiple zero-cost proxies can remain viable depending on the deployment constraints and computational budget.

\label{sec:accuracy_estimators}

\begin{figure*}[t]
    \centering
    \includegraphics[width=\textwidth]{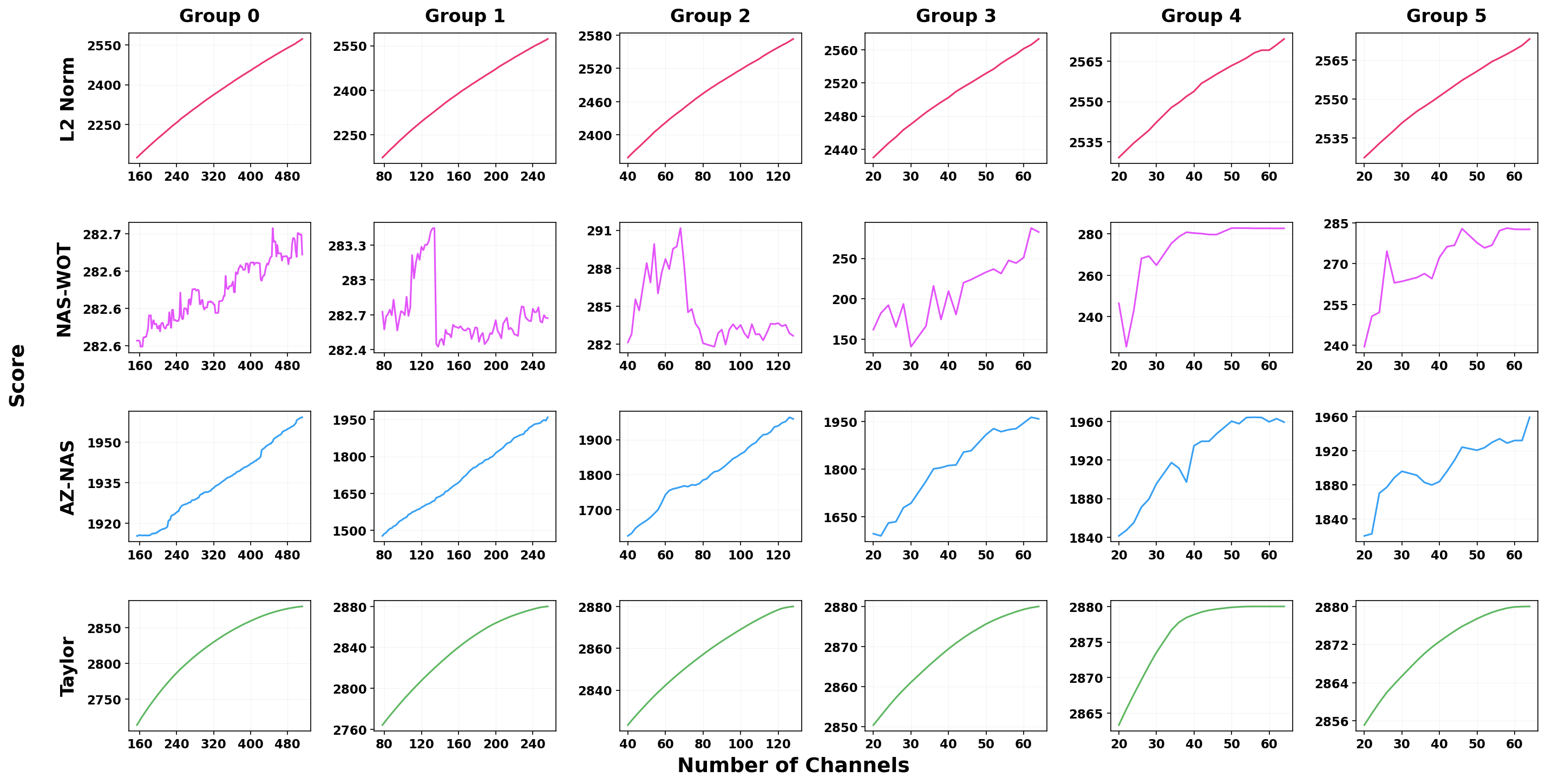}
    \caption{Group-wise score distributions for various zero-cost accuracy estimators (L2-Norm \cite{li2017pruning}, NWOT \cite{mellor2021neural} , AZ-NAS \cite{lee2024az}, and Taylor Importance \cite{molchanov2017pruningtaylor} across the ResNet-18 architecture. Each plot illustrates how the estimator's score changes with the number of channels.}
    \label{fig:metric_scores_comparison}
\end{figure*}

\subsubsection{Pruning Ratio}

We evaluate the effectiveness of different structured pruning methods in terms of multiply-accumulate (MAC) reduction on the Flowers102 dataset \cite{nilsback2008automated}. Table~\ref{tab:mac_flowers102} summarizes the results across multiple pruning ratios for both ResNet18 and MobileNetV2 architectures.

As shown in Table~\ref{tab:mac_flowers102}, we compare baseline models with pruning approaches including DepGraph~\cite{fang2023depgraph} and the proposed PruneNAS variants (one-shot and iterative). The pruning ratios range from mild compression (1.5$\times$) to aggressive reduction (9$\times$), allowing us to analyze performance trends under varying computational budgets.

Notably, at $6\times$ and $9\times$ MAC reduction on MobileNetV2, DepGraph suffers severe performance degradation, whereas iterative PruneNAS maintains robust performance, yielding an accuracy improvement of up to $\sim$20\%.

\subsubsection{Importance Criteria}
To evaluate the impact of different importance criteria within the PruneNAS framework, we compare Magnitude \cite{han2015learning}, Taylor~\cite{molchanov2017pruningtaylor}, Hessian~\cite{lecun1989optimal}, FPGM~\cite{fpgm}, and LAMP~\cite{lee2020layer}. We analyze their effect on the performance of the final pruned architecture using ResNet-18~\cite{he2016deep} on the Flowers102 dataset~\cite{nilsback2008automated}.

As shown in Table~\ref{tab:sparsity_accuracy}, the choice of importance criterion significantly affects the model's performance, particularly at aggressive compression ratios. LAMP achieves the highest retained accuracy overall, demonstrating exceptional robustness at higher compression levels. Taylor expansion shows strong performance at lower compression rates (e.g., $1.5\times$). Notably, despite its simplicity, the $\ell_{1}$ magnitude proxy exhibits remarkable consistency across all compression budgets, outperforming all other methods except LAMP at the extreme $9\times$ reduction.

\begin{figure}[t] 
    \centering
    \includegraphics[width=\linewidth]{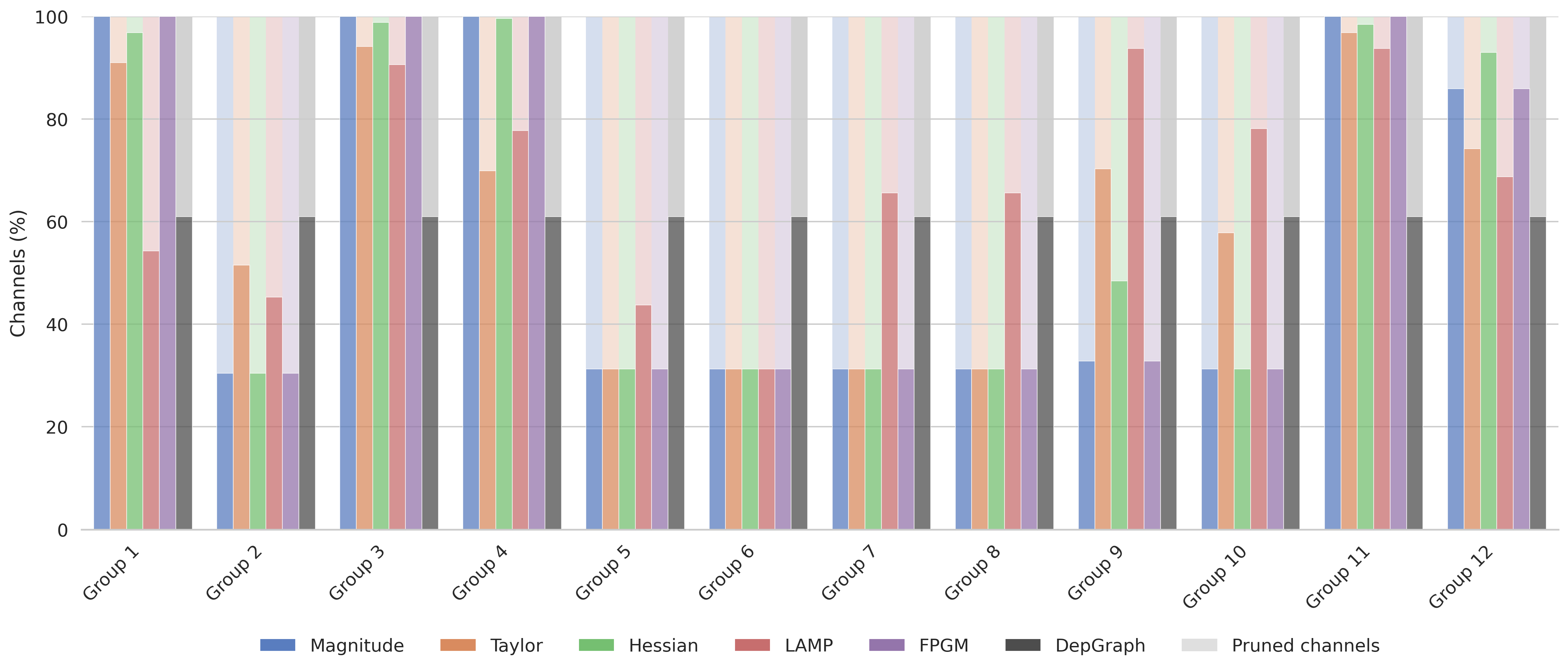} 
    \caption{Group-wise channel distribution at 3$\times$ Mac Reduction}
    \label{fig:ablation_3x}
\end{figure}


Figure \ref{fig:ablation_3x} visualizes the resulting group-wise channel distribution at 3$\times$ MAC reduction; additional compression ratios are shown in the Appendix. LAMP is fundamentally a magnitude-based variant, and at most compression ratios, their distributions are similar, but at 3$\times$, LAMP's layer-adaptive properties prevent the more aggressive middle-layer pruning. The structural bottlenecking could explain why it outperforms the standard magnitude proxy. At 9$\times$ compression ratio, almost all groups reach the 70\% prune ratio ceiling, but the resulting pruned models have distinct topologies. The precise group-wise distribution dictates the network's remaining representational capacity and explains why certain methods are superior to others.

\subsection{Hardware Benchmarking}


\begin{table}[h!]
\centering
\small 
\caption{Benchmarking of MobileNetV2 models on STM32H7S78-DK device (trained on Flowers102 datasets). Accuracy is reported on FP32 models; RAM and latency are measured on STM32H7~\cite{stmicroelectronics2024n6} using INT8 inference. $^{\ast}$ Indicates the use of external memory, which is slower than internal memory.}
\label{tab:benchmark_h7}
\begin{tabular}{l c c c c}
\hline
\textbf{Model} & \textbf{Params (M)} & \textbf{Top-1 Acc. (\%)} & \textbf{RAM (MB)} & \textbf{Latency (ms)} \\
\hline
\multicolumn{5}{c}{\textbf{Baseline}} \\
\hline 
\textbf{MobileNetV2} & 3.39 & 93.62 & 1.61 & 1304.00 \\
\hline 
\multicolumn{5}{c}{\textbf{DepGraph}} \\
\hline
{MAC/1.5 } & \textbf{1.59} & 90.86 & $1.34^{\ast}$ & \textbf{1034.00} \\
{MAC/3.0 } & \textbf{0.82} & 86.05 & $1.05^{\ast}$ & 542.60  \\
{MAC/6.0 } & \textbf{0.46} & 64.63 & 0.77 & 339.70 \\
\hline
\multicolumn{5}{c}{\textbf{PruneNAS}} \\
\hline 
{MAC/1.5 } & 2.10 & 90.86 & $\textbf{1.25}^\ast$ & 1072.00 \\
{MAC/3.0 } & 1.07 & \textbf{86.25} & \textbf{0.46} & \textbf{358.50} \\ 
{MAC/6.0 } & 0.53 & \textbf{84.35} & \textbf{0.44} & \textbf{202.80} \\
\hline 
\end{tabular}
\end{table}

We benchmarked pruned models generated using DepGraph~\cite{fang2023depgraph} and PruneNAS on edge AI hardware (STM32H7 board) featuring 620~KB internal RAM, 16~MB external RAM, 64~KB internal flash, and 128~MB external flash. It is important to note that accesses to external RAM and flash incur significantly higher latency compared to internal memory. As a result, models that minimize reliance on external memory demonstrate superior runtime performance.

Table~\ref{tab:benchmark_h7} shows that, under 3$\times$ and 6$\times$ MAC reduction budgets, PruneNAS achieves latency improvements of 3.64$\times$ and 6.42$\times$, respectively, compared to the baseline. In contrast, DepGraph achieves latency improvements of only 2.40$\times$ and 3.83$\times$ under the same budgets. Furthermore, PruneNAS models exhibit lower RAM usage than DepGraph models at both 3$\times$ and 6$\times$ reduction levels, which contributes to the additional speedup observed. On the other hand, DepGraph-pruned models tend to have smaller model sizes (i.e., lower flash requirements). This is because PruneNAS optimizes models under a MAC-based constraint, resulting in non-uniform pruning ratios across layers, whereas DepGraph applies a more uniform pruning strategy across layers for a given pruning budget. 

These results also underscore that MACs are an imperfect proxy for real-world efficiency. For instance, in a separate experiment on the STM32N6, an unpruned MobileNetV2 runs in 16.01\,ms, whereas a naive $1.5\times$ MAC-reduced model can be markedly slower (96.37\,ms) once its larger intermediate tensors spill into slower external memory. PruneNAS mitigates this by optimizing directly under hardware-aware constraints rather than proxy MACs alone.

\subsection{Limitations and Future Work}
While PruneNAS ensures strict constraint satisfaction and structural integrity by grouping dependent layers, our ILP formulation (Eq.~\ref{eq:ilp}) assumes independent pruning effects across groups. Furthermore, this work focuses on single-objective constraints and will transition to a unified joint-optimization solver in future work. Finally, we observe that under extreme compression regimes (e.g., $<0.15$G MACs on lightweight models like MobileNetV2), one-shot pruning can suffer from severe degradation and requires iterative fine-tuning to recover performance.
\section{Conclusion\label{sec:conclusion}}
In this paper, We presented PruneNAS, a structured pruning framework that efficiently generates candidate architectures via training-free tabulation followed by a global search. On ImageNet, it maintains competitive accuracy on ResNet-50 while nearly halving the MACs required by prior approaches. Furthermore, it improves accuracy by 1\% for ResNet, highlighting the importance of searching for optimal, non-uniform channel distributions.  At $9\times$ MACs reduction on flowers, it yields up to an improvement of $\sim 20\%$ over DepGraph on MobileNetV2. These savings translate directly to physical edge hardware, achieving up to a $6.42\times$ reduction in inference latency and significantly lower RAM utilization on an STM32H7 microcontroller. PruneNAS provides a fast, highly scalable, and hardware-validated solution for real-world model deployment.

\clearpage

\bibliography{egbib}

\begin{thebibliography}{63}
\providecommand{\natexlab}[1]{#1}
\providecommand{\url}[1]{\texttt{#1}}
\expandafter\ifx\csname urlstyle\endcsname\relax
  \providecommand{\doi}[1]{doi: #1}\else
  \providecommand{\doi}{doi: \begingroup \urlstyle{rm}\Url}\fi

\bibitem[Abdelfattah et~al.(2021)Abdelfattah, Mehrotra, Dudziak, and Lane]{abdelfattah2021zero}
Mohamed~S Abdelfattah, Abhinav Mehrotra, {\L}ukasz Dudziak, and Nicholas~D Lane.
\newblock Zero-cost proxies for lightweight nas.
\newblock \emph{arXiv preprint arXiv:2101.08134}, 2021.

\bibitem[Cai et~al.(2018)Cai, Zhu, and Han]{cai2018proxylessnas}
Han Cai, Ligeng Zhu, and Song Han.
\newblock Proxylessnas: Direct neural architecture search on target task and hardware.
\newblock \emph{arXiv preprint arXiv:1812.00332}, 2018.

\bibitem[Cai et~al.(2019)Cai, Gan, Wang, Zhang, and Han]{cai2019once}
Han Cai, Chuang Gan, Tianzhe Wang, Zhekai Zhang, and Song Han.
\newblock Once-for-all: Train one network and specialize it for efficient deployment.
\newblock \emph{arXiv preprint arXiv:1908.09791}, 2019.

\bibitem[Chen and Ran(2019)]{chen2019deep}
Jiasi Chen and Xukan Ran.
\newblock Deep learning with edge computing: A review.
\newblock \emph{Proceedings of the IEEE}, 107\penalty0 (8):\penalty0 1655--1674, 2019.

\bibitem[De~Ruyter et~al.(2026)De~Ruyter, Verbeke, and Hallez]{deruyter2026connas}
Gregory De~Ruyter, Mathias Verbeke, and Hans Hallez.
\newblock Constraint-guided hardware-aware neural architecture search through gradient modification.
\newblock In \emph{International Conference on Learning Representations (ICLR)}, 2026.
\newblock URL \url{https://openreview.net/forum?id=fcKMFIiNFQ}.

\bibitem[Deng et~al.(2009)Deng, Dong, Socher, Li, Li, and Fei-Fei]{deng2009imagenet}
Jia Deng, Wei Dong, Richard Socher, Li-Jia Li, Kai Li, and Li~Fei-Fei.
\newblock Imagenet: A large-scale hierarchical image database.
\newblock In \emph{2009 IEEE conference on computer vision and pattern recognition}, pages 248--255. Ieee, 2009.

\bibitem[Ding et~al.(2019{\natexlab{a}})Ding, Ding, Guo, and Han]{aofp}
Xiaohan Ding, Guiguang Ding, Yuchen Guo, and Jungong Han.
\newblock Accelerated filter pruning with structural redundancy reduction.
\newblock In \emph{CVPR}, 2019{\natexlab{a}}.

\bibitem[Ding et~al.(2019{\natexlab{b}})Ding, Ding, Guo, and Han]{ding2019centripetal}
Xiaohan Ding, Guiguang Ding, Yuchen Guo, and Jungong Han.
\newblock Centripetal sgd for pruning very deep convolutional networks with complicated structure.
\newblock In \emph{Proceedings of the IEEE/CVF conference on computer vision and pattern recognition}, pages 4943--4953, 2019{\natexlab{b}}.

\bibitem[Ding et~al.(2022)Ding, Wu, Huang, Tang, Wu, Yang, Zhu, and Zhuang]{ding2022nap}
Yadong Ding, Yu~Wu, Chengyue Huang, Siliang Tang, Fei Wu, Yi~Yang, Wenwu Zhu, and Yueting Zhuang.
\newblock Nap: Neural architecture search with pruning.
\newblock \emph{Neurocomputing}, 477:\penalty0 85--95, 2022.

\bibitem[Dong and Yang(2019{\natexlab{a}})]{dong2019network}
Xuanyi Dong and Yi~Yang.
\newblock Network pruning via transformable architecture search.
\newblock \emph{Advances in neural information processing systems}, 32, 2019{\natexlab{a}}.

\bibitem[Dong and Yang(2019{\natexlab{b}})]{tas}
Xuanyi Dong and Yi~Yang.
\newblock Task-aware network compression via joint architecture search and pruning.
\newblock In \emph{CVPR}, 2019{\natexlab{b}}.

\bibitem[Dunning et~al.(2011)Dunning, Mitchell, and O'Sullivan]{dunning2011pulp}
Iain Dunning, Stuart Mitchell, and Michael O'Sullivan.
\newblock Pulp: a linear programming toolkit for python.
\newblock \emph{Optimization Online}, 2011.
\newblock URL \url{https://optimization-online.org/2011/09/3178/}.

\bibitem[Fang et~al.(2023)Fang, Ma, Song, Mi, and Wang]{fang2023depgraph}
Gongfan Fang, Xinyin Ma, Mingli Song, Michael~Bi Mi, and Xinchao Wang.
\newblock Depgraph: Towards any structural pruning.
\newblock In \emph{Proceedings of the IEEE/CVF conference on computer vision and pattern recognition}, pages 16091--16101, 2023.

\bibitem[Han et~al.(2015)Han, Pool, Tran, and Dally]{han2015learning}
Song Han, Jeff Pool, John Tran, and William~J. Dally.
\newblock Learning both weights and connections for efficient neural network.
\newblock In \emph{Advances in Neural Information Processing Systems (NeurIPS)}, 2015.

\bibitem[He et~al.(2016)He, Zhang, Ren, and Sun]{he2016deep}
Kaiming He, Xiangyu Zhang, Shaoqing Ren, and Jian Sun.
\newblock Deep residual learning for image recognition.
\newblock In \emph{Proceedings of the IEEE conference on computer vision and pattern recognition}, pages 770--778, 2016.

\bibitem[He et~al.(2018{\natexlab{a}})He, Kang, Dong, Fu, and Yang]{sfp}
Yang He, Guoliang Kang, Xuanyi Dong, Yanwei Fu, and Yi~Yang.
\newblock Soft filter pruning for accelerating deep convolutional neural networks.
\newblock In \emph{IJCAI}, 2018{\natexlab{a}}.

\bibitem[He et~al.(2019)He, Liu, Wang, Hu, and Yang]{fpgm}
Yang He, Ping Liu, Ziwei Wang, Zhilan Hu, and Yi~Yang.
\newblock Filter pruning via geometric median for deep convolutional neural networks acceleration.
\newblock In \emph{CVPR}, 2019.

\bibitem[He et~al.(2018{\natexlab{b}})He, Lin, Liu, Wang, Li, and Han]{he2018amc}
Yihui He, Ji~Lin, Zhijian Liu, Hanrui Wang, Li-Jia Li, and Song Han.
\newblock Amc: Automl for model compression and acceleration on mobile devices.
\newblock In \emph{Proceedings of the European conference on computer vision (ECCV)}, pages 784--800, 2018{\natexlab{b}}.

\bibitem[He et~al.(2017)]{he2017channel}
Yihui He et~al.
\newblock Channel pruning for accelerating very deep neural networks.
\newblock In \emph{ICCV}, 2017.

\bibitem[Hou et~al.(2022)Hou, Qin, Sun, Ma, Yuan, Xu, Chen, Jin, Xie, and Kung]{hou2022chex}
Zejiang Hou, Minghai Qin, Fei Sun, Xiaolong Ma, Kun Yuan, Yi~Xu, Yen-Kuang Chen, Rong Jin, Yuan Xie, and Sun-Yuan Kung.
\newblock Chex: Channel exploration for cnn model compression.
\newblock In \emph{Proceedings of the IEEE/CVF Conference on Computer Vision and Pattern Recognition}, pages 12287--12298, 2022.

\bibitem[Krizhevsky(2009)]{krizhevsky2009learning}
Alex Krizhevsky.
\newblock Learning multiple layers of features from tiny images.
\newblock Technical report, University of Toronto, 2009.
\newblock Technical Report.

\bibitem[LeCun et~al.(1989)LeCun, Denker, and Solla]{lecun1989optimal}
Yann LeCun, John Denker, and Sara Solla.
\newblock Optimal brain damage.
\newblock \emph{Advances in neural information processing systems}, 2, 1989.

\bibitem[Lee et~al.(2020)Lee, Park, Mo, Ahn, and Shin]{lee2020layer}
Jaeho Lee, Sejun Park, Sangwoo Mo, Sungsoo Ahn, and Jinwoo Shin.
\newblock Layer-adaptive sparsity for the magnitude-based pruning.
\newblock \emph{arXiv preprint arXiv:2010.07611}, 2020.

\bibitem[Lee and Ham(2024)]{lee2024az}
Junghyup Lee and Bumsub Ham.
\newblock Az-nas: Assembling zero-cost proxies for network architecture search.
\newblock In \emph{Proceedings of the IEEE/CVF conference on computer vision and pattern recognition}, pages 5893--5903, 2024.

\bibitem[Lee et~al.(2019)Lee, Ajanthan, and Torr]{lee2018snip}
Namhoon Lee, Thalaiyasingam Ajanthan, and Philip~HS Torr.
\newblock Snip: Single-shot network pruning based on connection sensitivity.
\newblock In \emph{ICLR}, 2019.

\bibitem[Lemaire et~al.(2019)Lemaire, Achkar, and Jodoin]{lemaire2018structured}
Carl Lemaire, Andrew Achkar, and Pierre{-}Marc Jodoin.
\newblock Structured pruning of neural networks with budget-aware regularization.
\newblock In \emph{Proceedings of the IEEE/CVF Conference on Computer Vision and Pattern Recognition (CVPR)}, pages 9108--9116, 2019.

\bibitem[Li et~al.(2023)Li, Yang, Bhardwaj, and Marculescu]{li2023zico}
Guihong Li, Yuedong Yang, Kartikeya Bhardwaj, and Radu Marculescu.
\newblock Zico: Zero-shot nas via inverse coefficient of variation on gradients.
\newblock \emph{arXiv preprint arXiv:2301.11300}, 2023.

\bibitem[Li et~al.(2017{\natexlab{a}})Li, Kadav, Durdanovic, Samet, and Graf]{li2017pruning}
Hao Li, Asim Kadav, Igor Durdanovic, Hanan Samet, and Hans~Peter Graf.
\newblock Pruning filters for efficient convnets.
\newblock In \emph{International Conference on Learning Representations}, 2017{\natexlab{a}}.

\bibitem[Li et~al.(2017{\natexlab{b}})]{li2016pruning}
Hao Li et~al.
\newblock Pruning filters for efficient convnets.
\newblock In \emph{ICLR}, 2017{\natexlab{b}}.

\bibitem[Li et~al.(2022)Li, Zhao, Yuan, Lin, Wang, and Chen]{li2022pas}
Yanyu Li, Pu~Zhao, Geng Yuan, Xue Lin, Yanzhi Wang, and Xin Chen.
\newblock Pruning-as-search: Efficient neural architecture search via channel pruning and structural reparameterization.
\newblock In \emph{IJCAI}, 2022.

\bibitem[Lin et~al.(2018)Lin, Ji, et~al.]{gfp}
Shaohui Lin, Rongrong Ji, et~al.
\newblock Global filter pruning for accelerating deep convolutional neural networks.
\newblock In \emph{IJCAI}, 2018.

\bibitem[Lin et~al.(2024)Lin, Endo, Lee, and Kamijo]{lin2024ttnas}
Yiqi Lin, Yuki Endo, Jinho Lee, and Shunsuke Kamijo.
\newblock Neural architecture search via trainless pruning algorithm: A bayesian evaluation of a network with multiple indicators.
\newblock \emph{Electronics}, 13\penalty0 (22):\penalty0 4547, 2024.
\newblock \doi{10.3390/electronics13224547}.

\bibitem[Liu et~al.(2019{\natexlab{a}})]{liu2019darts}
Hanxiao Liu et~al.
\newblock Darts: Differentiable architecture search.
\newblock In \emph{ICLR}, 2019{\natexlab{a}}.

\bibitem[Liu et~al.(2021)Liu, Zhang, Kuang, Zhou, Xue, Wang, Chen, Yang, Liao, and Zhang]{liu2021group}
Liyang Liu, Shilong Zhang, Zhanghui Kuang, Aojun Zhou, Jing-Hao Xue, Xinjiang Wang, Yimin Chen, Wenming Yang, Qingmin Liao, and Wayne Zhang.
\newblock Group fisher pruning for practical network compression.
\newblock In \emph{International Conference on Machine Learning}, pages 7021--7032. PMLR, 2021.

\bibitem[Liu et~al.(2022)Liu, Chen, Chen, Shen, Mocanu, Wang, and Pechenizkiy]{liu2022unreasonable}
Shiwei Liu, Tianlong Chen, Xiaohan Chen, Li~Shen, Decebal~Constantin Mocanu, Zhangyang Wang, and Mykola Pechenizkiy.
\newblock The unreasonable effectiveness of random pruning: Return of the most naive baseline for sparse training.
\newblock \emph{arXiv preprint arXiv:2202.02643}, 2022.

\bibitem[Liu et~al.(2019{\natexlab{b}})]{meta}
Zechun Liu et~al.
\newblock Metapruning: Meta learning for automatic neural network channel pruning.
\newblock In \emph{ICCV}, 2019{\natexlab{b}}.

\bibitem[Luo et~al.(2017)Luo, Wu, and Lin]{thinet}
Jian-Hao Luo, Jianxin Wu, and Weiyao Lin.
\newblock Thinet: A filter level pruning method for deep neural network compression.
\newblock In \emph{ICCV}, 2017.

\bibitem[Lv et~al.(2026)Lv, Zhou, Tian, Ye, Feng, and Lv]{lv2026hypernas}
Jindi Lv, Yuhao Zhou, Yuxin Tian, Qing Ye, Wentao Feng, and Jiancheng Lv.
\newblock Hypernas: Enhancing architecture representation for nas predictor via hypernetwork.
\newblock In \emph{Proceedings of the IEEE/CVF Conference on Computer Vision and Pattern Recognition (CVPR)}, 2026.

\bibitem[Mellor et~al.(2021)Mellor, Turner, Storkey, and Crowley]{mellor2021neural}
Joe Mellor, Jack Turner, Amos Storkey, and Elliot~J Crowley.
\newblock Neural architecture search without training.
\newblock In \emph{International conference on machine learning}, pages 7588--7598. PMLR, 2021.

\bibitem[Menghani(2023)]{menghani2023efficient}
Gaurav Menghani.
\newblock Efficient deep learning: A survey on making deep learning models smaller, faster, and better.
\newblock \emph{ACM Computing Surveys}, 55\penalty0 (12):\penalty0 1--37, 2023.

\bibitem[Molchanov et~al.(2017)Molchanov, Tyree, Karras, Aila, and Kautz]{molchanov2017pruningtaylor}
Pavlo Molchanov, Stephen Tyree, Tero Karras, Timo Aila, and Jan Kautz.
\newblock Pruning convolutional neural networks for resource efficient inference.
\newblock In \emph{International Conference on Learning Representations (ICLR)}, 2017.

\bibitem[Molchanov et~al.(2019)Molchanov, Mallya, Tyree, Frosio, and Kautz]{molchanov2019importance}
Pavlo Molchanov, Arun Mallya, Stephen Tyree, Iuri Frosio, and Jan Kautz.
\newblock Importance estimation for neural network pruning.
\newblock In \emph{Proceedings of the IEEE/CVF conference on computer vision and pattern recognition}, pages 11264--11272, 2019.

\bibitem[Molchanov et~al.(2022)Molchanov, Hall, Yin, Kautz, Fusi, and Vahdat]{molchanov2022lana}
Pavlo Molchanov, Jimmy Hall, Hongxu Yin, Jan Kautz, Nicol{\`o} Fusi, and Arash Vahdat.
\newblock Lana: Latency aware network acceleration.
\newblock In \emph{European Conference on Computer Vision (ECCV)}, Lecture Notes in Computer Science, pages 137--156. Springer, 2022.

\bibitem[Nilsback and Zisserman(2008)]{nilsback2008automated}
Maria-Elena Nilsback and Andrew Zisserman.
\newblock Automated flower classification over a large number of classes.
\newblock In \emph{2008 Sixth Indian conference on computer vision, graphics \& image processing}, pages 722--729. IEEE, 2008.

\bibitem[Peng et~al.(2018)Peng, Wu, Chen, et~al.]{ccp}
Hanting Peng, Jing Wu, Shuxin Chen, et~al.
\newblock Channel pruning via conditional mutual information.
\newblock In \emph{ECCV}, 2018.

\bibitem[Sandler et~al.(2018)Sandler, Howard, Zhu, Zhmoginov, and Chen]{sandler2018mobilenetv2}
Mark Sandler, Andrew Howard, Menglong Zhu, Andrey Zhmoginov, and Liang-Chieh Chen.
\newblock Mobilenetv2: Inverted residuals and linear bottlenecks.
\newblock In \emph{Proceedings of the IEEE Conference on Computer Vision and Pattern Recognition}, pages 4510--4520, 2018.

\bibitem[{STMicroelectronics}(2024)]{stmicroelectronics2024n6}
{STMicroelectronics}.
\newblock Stm32n6 ai npu platform for next-generation embedded vision, 2024.
\newblock \url{https://www.st.com/en/microcontrollers-microprocessors/stm32n6.html}.

\bibitem[Tanaka et~al.(2020)Tanaka, Kunin, Yamins, and Ganguli]{tanaka2020pruning}
Hidenori Tanaka, Daniel Kunin, Daniel~L Yamins, and Surya Ganguli.
\newblock Pruning neural networks without any data by iteratively conserving synaptic flow.
\newblock \emph{Advances in neural information processing systems}, 33:\penalty0 6377--6389, 2020.

\bibitem[Theis et~al.(2018)Theis, Korshunova, Tejani, and Husz{\'a}r]{theis2018faster}
Lucas Theis, Iryna Korshunova, Alykhan Tejani, and Ferenc Husz{\'a}r.
\newblock Faster gaze prediction with dense networks and fisher pruning.
\newblock \emph{arXiv preprint arXiv:1801.05787}, 2018.

\bibitem[{Ultralytics}(2023)]{yolov8_ultralytics}
{Ultralytics}.
\newblock Yolov8: Ultralytics yolo.
\newblock \url{https://github.com/ultralytics/ultralytics}, 2023.
\newblock Accessed: 2026-04-29.

\bibitem[Wang et~al.(2020{\natexlab{a}})Wang, Zhang, and Grosse]{Wang2020Picking}
Chaoqi Wang, Guodong Zhang, and Roger Grosse.
\newblock Picking winning tickets before training by preserving gradient flow.
\newblock In \emph{International Conference on Learning Representations}, 2020{\natexlab{a}}.
\newblock URL \url{https://openreview.net/forum?id=SkgsACVKPH}.

\bibitem[Wang et~al.(2021)Wang, Qin, Zhang, and Fu]{greg}
Huan Wang, Can Qin, Yulun Zhang, and Yun Fu.
\newblock Neural pruning via growing regularization.
\newblock In \emph{International Conference on Learning Representations (ICLR)}, 2021.
\newblock URL \url{https://arxiv.org/abs/2012.09243}.

\bibitem[Wang et~al.(2020{\natexlab{b}})Wang, Wang, Cai, Lin, Liu, Wang, Lin, and Han]{wang2020apq}
Tianzhe Wang, Kuan Wang, Han Cai, Ji~Lin, Zhijian Liu, Hanrui Wang, Yujun Lin, and Song Han.
\newblock Apq: Joint search for network architecture, pruning and quantization policy.
\newblock In \emph{Proceedings of the IEEE/CVF Conference on Computer Vision and Pattern Recognition (CVPR)}, pages 2075--2084, 2020{\natexlab{b}}.
\newblock \doi{10.1109/CVPR42600.2020.00215}.

\bibitem[Wen et~al.(2016)]{wen2016learning}
Wei Wen et~al.
\newblock Learning structured sparsity in deep neural networks.
\newblock In \emph{NeurIPS}, 2016.

\bibitem[Wightman et~al.(2021)Wightman, Touvron, and J{\'e}gou]{wightman2021resnet}
Ross Wightman, Hugo Touvron, and Herv{\'e} J{\'e}gou.
\newblock Resnet strikes back: An improved training procedure in timm.
\newblock \emph{arXiv preprint arXiv:2110.00476}, 2021.

\bibitem[Wu et~al.(2019)Wu, Dai, Zhang, Wang, Sun, Wu, Tian, Vajda, Jia, and Keutzer]{wu2019fbnet}
Bichen Wu, Xiaoliang Dai, Peizhao Zhang, Yanghan Wang, Fei Sun, Yiming Wu, Yuandong Tian, Peter Vajda, Yangqing Jia, and Kurt Keutzer.
\newblock Fbnet: Hardware-aware efficient convnet design via differentiable neural architecture search.
\newblock In \emph{Proceedings of the IEEE/CVF conference on computer vision and pattern recognition}, pages 10734--10742, 2019.

\bibitem[Yang et~al.(2025)Yang, Liu, Wang, Wu, Chen, and Ma]{yang2025patnas}
Jiechao Yang, Yong Liu, Wei Wang, Haoran Wu, Zhiyuan Chen, and Xibo Ma.
\newblock Patnas: A path-based training-free neural architecture search.
\newblock \emph{IEEE Transactions on Pattern Analysis and Machine Intelligence}, 47\penalty0 (11):\penalty0 9876--9893, 2025.
\newblock \doi{10.1109/TPAMI.2024.3498035}.

\bibitem[Yang et~al.(2018)Yang, Howard, et~al.]{netadapt}
Tien-Ju Yang, Andrew Howard, et~al.
\newblock Netadapt: Platform-aware neural network adaptation for mobile applications.
\newblock In \emph{ECCV}, 2018.

\bibitem[You et~al.(2019)You, Yan, Ye, Ma, and Wang]{you2019gate}
Zhonghui You, Kun Yan, Jinmian Ye, Meng Ma, and Ping Wang.
\newblock Gate decorator: Global filter pruning method for accelerating deep convolutional neural networks.
\newblock \emph{Advances in neural information processing systems}, 32, 2019.

\bibitem[Yu and Huang(2020)]{autoslim}
Jiahui Yu and Thomas Huang.
\newblock Autoslim: Towards one-shot architecture search for channel numbers.
\newblock In \emph{ICLR}, 2020.

\bibitem[Yu and Huang(2019)]{yu2019universally}
Jiahui Yu and Thomas~S Huang.
\newblock Universally slimmable networks and improved training techniques.
\newblock In \emph{Proceedings of the IEEE/CVF international conference on computer vision}, pages 1803--1811, 2019.

\bibitem[Yu et~al.(2019)Yu, Yang, Xu, Yang, and Huang]{slimmable}
Jiahui Yu, Linjie Yang, Ning Xu, Jianchao Yang, and Thomas Huang.
\newblock Slimmable neural networks.
\newblock In \emph{ICLR}, 2019.

\bibitem[Yu et~al.(2022)Yu, Yao, Gholami, Dong, Kim, Mahoney, and Keutzer]{yu2022hessian}
Shixing Yu, Zhewei Yao, Amir Gholami, Zhen Dong, Sehoon Kim, Michael~W Mahoney, and Kurt Keutzer.
\newblock Hessian-aware pruning and optimal neural implant.
\newblock In \emph{Proceedings of the IEEE/CVF Winter Conference on Applications of Computer Vision}, pages 3880--3891, 2022.

\end{thebibliography}
\newpage
\section{Appendix}
\label{sec:appendix_distributions}

\begin{figure}[h] 
    \centering
    \includegraphics[width=\linewidth]{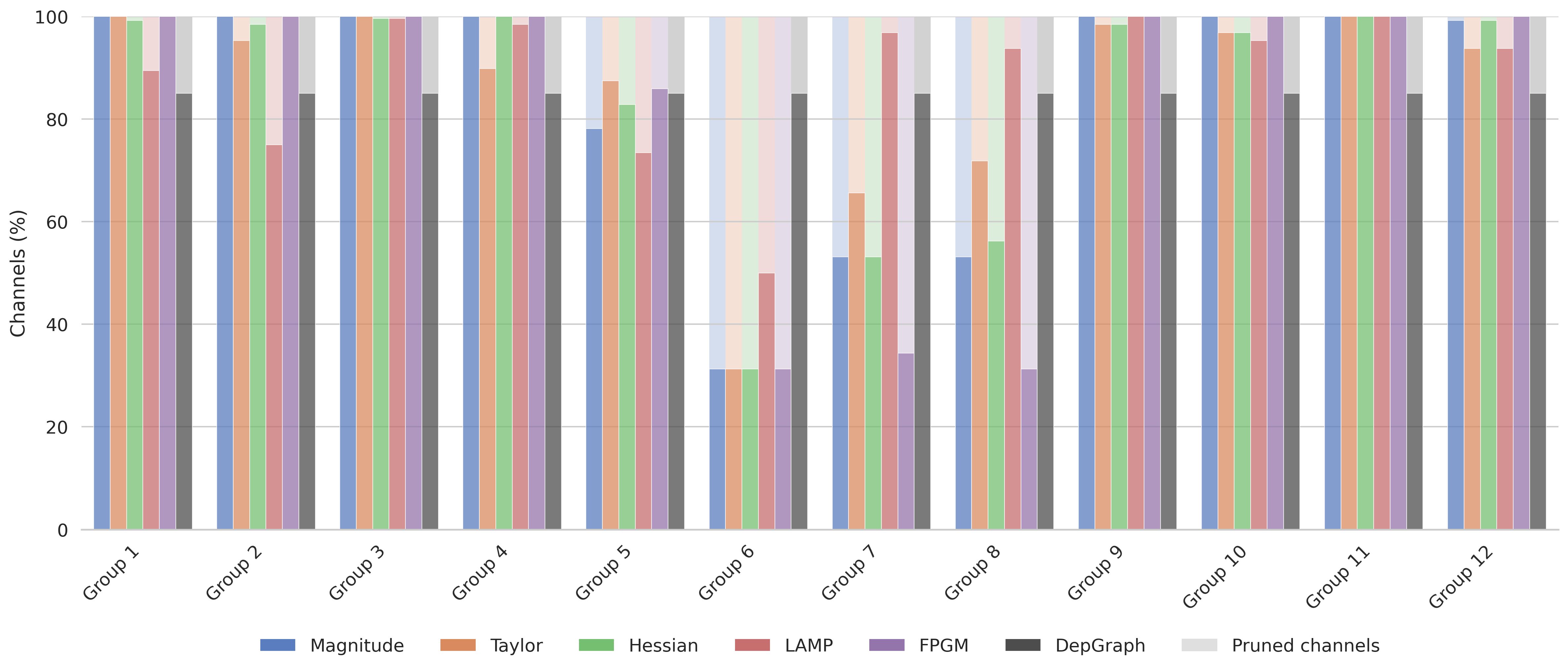} 
    \caption{Group-wise channel distribution at 1.5$\times$ Mac Reduction}
    \label{fig:ablation_15x}
\end{figure}

\begin{figure}[h] 
    \centering
    \includegraphics[width=\linewidth]{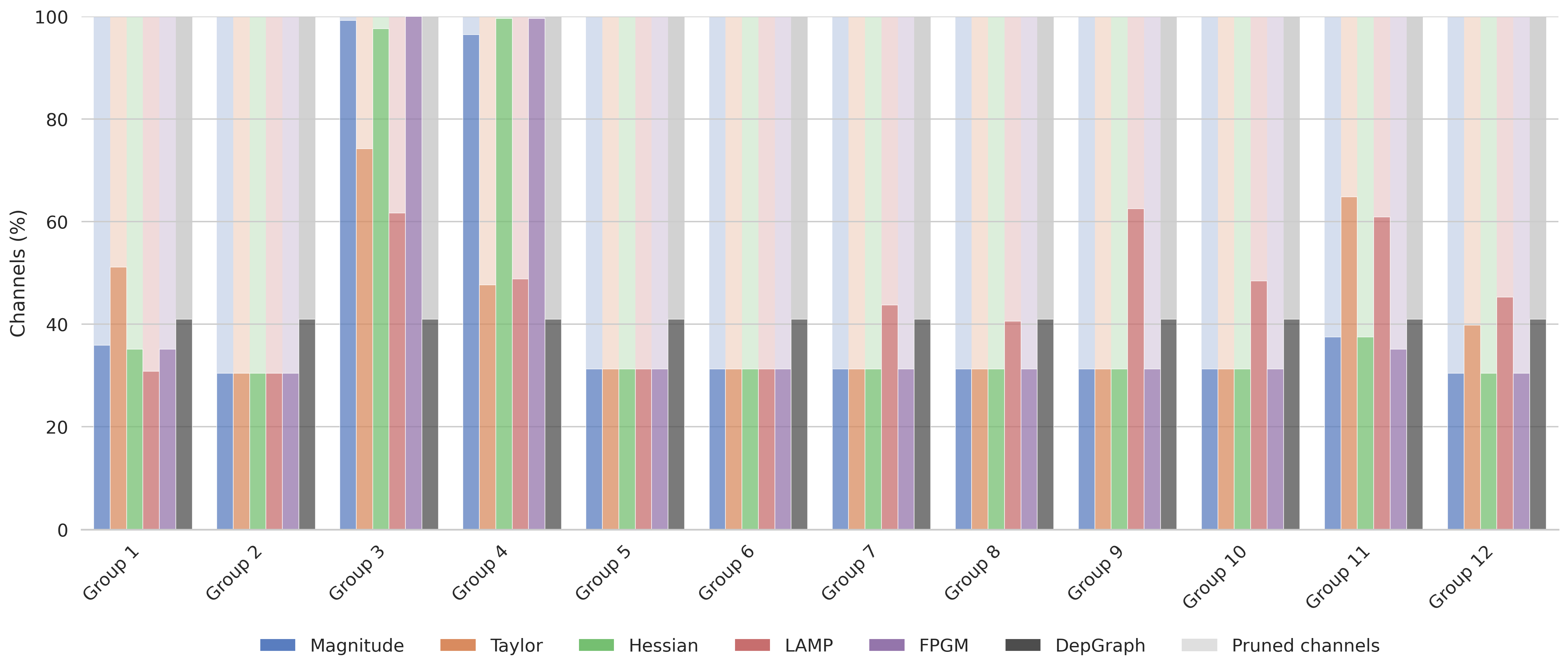} 
    \caption{Group-wise channel distribution at 6$\times$ Mac Reduction}
    \label{fig:ablation_6x}
\end{figure}

\begin{figure}[h] 
    \centering
    \includegraphics[width=\linewidth]{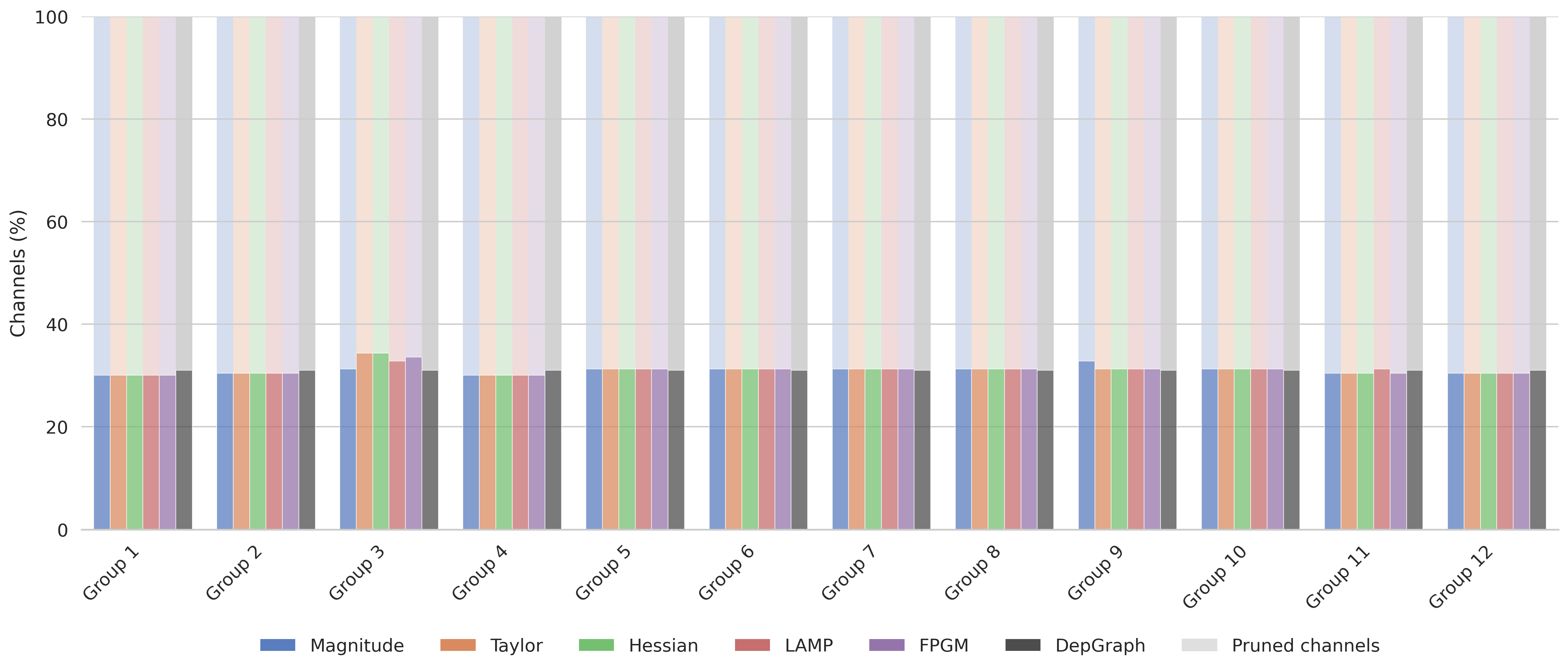} 
    \caption{Group-wise channel distribution at 9$\times$ Mac Reduction}
    \label{fig:ablation_9x}
\end{figure}

\begin{table}[h!]
\centering
\small
\caption{Summary of channel importance metrics used for structured pruning.}
\label{tab:pruning_metrics}
\begin{tabular}{p{3.0cm} p{4.0cm} p{4.5cm}}
\toprule
\textbf{Method} & \textbf{Description} & \textbf{Importance Score} \\
\midrule

Weight Magnitude \cite{han2015learning}  &
Assumes that weights with smaller absolute values contribute less to the network output. &
$\displaystyle S_{\text{mag}}(c) = \sum_{w \in W_c} |w|$ \\

\midrule

First-Order Taylor \cite{molchanov2017pruningtaylor} &
Estimates change in loss $\mathcal{L}$ when a channel is removed using first-order gradient information. &
$\displaystyle S_{\text{taylor}}(c) = \sum_{w \in W_c} \left| w \frac{\partial \mathcal{L}}{\partial w} \right|$ \\

\midrule

Hessian Approximation \cite{lecun1989optimal} &
Captures curvature of the loss using a diagonal approximation of the Hessian (empirical Fisher). &
$\displaystyle S_{\text{hessian}}(c) = \sum_{w \in W_c} w^2 \left( \frac{\partial \mathcal{L}}{\partial w} \right)^2$ \\

\midrule

LAMP \cite{lee2020layer} &
Normalizes magnitude scores to produce globally comparable importance across layers. &
$\displaystyle S_{\text{lamp}}(c_u) = \frac{S_{\text{mag}}(c_u)}{\sum_{v \ge u} S_{\text{mag}}(c_v)}$ \\

\midrule

FPGM \cite{fpgm} &
Identifies redundant channels via geometric similarity (Euclidean distance) within a group $\mathcal{G}$. &
$\displaystyle S_{\text{fpgm}}(c) = \sum_{c' \in \mathcal{G}} \left\| W_c - W_{c'} \right\|_2$ \\

\bottomrule
\end{tabular}
\end{table}

\begin{table}[h]
\centering
\caption{Comparison of top-1 accuracy across various estimators at a fixed 9$\times$ MAC reduction, evaluated using ResNet-18 (dense model top-1 accuracy 91.87) on the Flowers102 dataset \cite{nilsback2008automated}.}
\label{tab:estimators_ablation}
\begin{tabular}{lc}
\toprule
\multirow{2}{*}{\textbf{Estimator}} &
\multicolumn{1}{c}{\textbf{MAC Reduction}} \\
\cmidrule(lr){2-2}
 & \textbf{9$\times$} \\
\midrule
L2-Norm \cite{li2017pruning} & 82.35 \\
NWOT \cite{mellor2021neural} & 82.61 \\
AZ-NAS \cite{lee2024az} & 81.80 \\
Taylor \cite{molchanov2017pruningtaylor} & \textbf{82.79} \\

\bottomrule
\end{tabular}
\end{table}

\begin{table}[t]
\centering
\caption{Accuracy--MAC reduction trade-offs on the CIFAR100 for ResNet-18 model~\cite{krizhevsky2009learning} dataset. ``Pruned'' denotes the accuracy immediately after pruning and before fine-tuning.}
\label{tab:mac_cifar100}
\begin{tabular}{lccccc}
\toprule
\multirow{2}{*}{\textbf{Method}} & \multicolumn{5}{c}{\textbf{MAC Reduction}} \\
\cmidrule(lr){2-6}
 & 1.0$\times$ & 1.5$\times$ & 3.0$\times$ & 6.0$\times$ & 9.0$\times$ \\
\midrule

Pruned & 77.48 & 50.51 & 7.83 & 2.30 & 1.46 \\

DepGraph~\cite{fang2023depgraph}
& 77.48 & 76.90 & 75.99 & 74.38 & 73.06 \\
& & (-0.58) & (-1.49) & (-3.10) & (-4.42) \\

\textbf{PruneNAS}
& 76.93 & 77.02 & 76.34 & 74.57 & 73.10 \\
& & \textbf{(+0.44)} & \textbf{(-0.59)} & \textbf{(-2.36)} & \textbf{(-3.83)} \\

\bottomrule
\end{tabular}
\end{table}

\begin{table}[h]
\centering
\caption{Tradeoff between granularity and optimization quality for ResNet-18 on the CIFAR-100~\cite{krizhevsky2009learning} dataset. The model is pruned to approximately one-third of its MAC, and Top-1 accuracy is reported after fine-tuning.}
\label{tab:granularity}
\begin{tabular}{lccc}
\toprule
\textbf{Granularity} & \textbf{Top-1 Accuracy (\%)} & \textbf{MAC Reduction} & \textbf{Tabulation Time (s)} \\
\midrule
2  & 75.61 & 2.994$\times$ & 135.32 \\
4  & 75.90 & 3.021$\times$ & 69.64 \\
8  & 76.34 & 3.001$\times$ & 36.11 \\
16 & 75.88 & 3.007$\times$ & 19.92 \\
32 & 76.18 & 3.012$\times$ & 11.27 \\
\bottomrule
\end{tabular}
\end{table}
\begin{table}[t]
\centering
\caption{Per-group pruning-cap sensitivity for ResNet-18 on CIFAR-100
at $9.01\times$ compression ($557.9\!\to\!61.9$\,MMACs; baseline 76.91\%).}
\label{tab:cap_sensitivity_9x}
\begin{tabular}{ccc}
\toprule
Max Pruning Percentage & Top-1 Acc (\%) & $\Delta$ vs.\ baseline \\
\midrule
$90\%$ & 70.43 & $-6.48$ \\
$80\%$ & 72.41 & $-4.50$ \\
$70\%$ & \textbf{73.10} & \textbf{$-3.81$} \\
\bottomrule
\end{tabular}
\end{table}

\end{document}